\documentclass[a4paper, 10pt, conference]{ieeeconf}

\IEEEoverridecommandlockouts  
\usepackage{epsfig}
\usepackage{graphicx}
\usepackage{amsmath}
\usepackage{amssymb}
\usepackage{comment}
\usepackage{subfigure}
\usepackage{float}

\usepackage{mars}

\begin{document}
\title{\LARGE \bf CP$^+$: Camera Poses Augmentation with Large-scale LiDAR Maps
}

\author{Jiadi Cui$^{1}$ and Sören Schwertfeger$^{2}$
\thanks{$^{1}$ J. Cui is with the School of Information Science and Technology, ShanghaiTech University, Shanghai 201210, China; the University of Chinese Academy of Sciences, Beijing 100049, China; the Shanghai Institute of Microsystem and Information Technology, Chinese Academy of Sciences, Shanghai 200050, China. \tt\small e-mail: cuijd@shanghaitech.edu.cn}
\thanks{$^{2}$ S. Schwertfeger is with the School of Information Science and Technology, ShanghaiTech University, Shanghai 201210, China. \tt\small e-mail: soerensch@shanghaitech.edu.cn}
}

%
%


\marsPublishedIn{Published with:} 	

\marsVenue{IEEE International Conference on Real-time Computing and Robotics (RCAR) 2022}

\marsYear{2022}

\marsPlainAutors{Jiadi Cui and Sören Schwertfeger}


\marsMakeCitation{CP$^+$: Camera Poses Augmentation with Large-scale LiDAR Maps}{IEEE Press}

\marsDOI{\url{https://doi.org/10.1109/RCAR54675.2022.9872176}}

\marsIEEE{}


\makeMARStitle

\maketitle
\begin{abstract} 
Large-scale colored point clouds have many advantages in navigation or scene display. Relying on cameras and LiDARs, which are now widely used in reconstruction tasks, it is possible to obtain such colored point clouds.
However, the information from these two kinds of sensors is not well fused in many existing frameworks, resulting in poor colorization results, thus resulting in inaccurate camera poses and damaged point colorization results. We propose a novel framework called Camera Pose Augmentation (CP$^+$) to improve the camera poses and align them directly with the LiDAR-based point cloud. Initial coarse camera poses are given by LiDAR-Inertial or LiDAR-Inertial-Visual Odometry with approximate extrinsic parameters and time synchronization. The key steps to improve the alignment of the images consist of selecting a point cloud corresponding to a region of interest in each camera view, extracting reliable edge features from this point cloud, and deriving 2D-3D line correspondences which are used towards iterative minimization of the re-projection error.
\end{abstract}

\section{Introduction}

Point clouds and optical images are useful for scene understanding, and cameras, LiDARs (Light Detection And Ranging) and Inertial Measurement Units (IMU) are widely employed in the underlying reconstruction and modeling tasks. However, combining the two modalities in order to reconstruct an RGB-colored point cloud is a highly challenging problem. Many applications utilize colorization or registration of 2D and 3D information, so it is very useful to get a better point cloud coloring.


A simple approach will just calibrate the extrinsic poses of all sensors \cite{chen2019heterogeneous} and utilize hardware synchronization to match the rigidly mounted sensors \cite{chen2020advanced}. But in real systems often this calibration or hardware synchronization is not available or the error is too big, such that more advanced methods need to be employed to generate good point cloud coloring.

A popular approach to  point cloud coloring are perspective-n-Point based (PnP-based) algorithms \cite{haralick1994review, lepetit2009epnp, kneip2014upnp}. They can handle the 2D-3D registration problem well, when there are several reliable correspondences. However, finding these correspondences is difficult for most real situations, and thus there are many methods that only use the same sensor for these tasks, even localization and mapping, which reveals a modality gap on mutual information registration which is hard to overcome. 

An intuitive idea is to construct a small 3D map with a series of images using a Structure-from-Motion (SFM) \cite{schonberger2016structure} approach, and then to match the map into the prior LiDAR map by Iterative Closest Point (ICP)-like methods with scale. This idea is computationally intensive. In addition, this approach needs the feature descriptors or intensities, which may be inconsistent under changing circumstances. Since directly registering two different types of sensor information is error-prone, we will aim at geometry-feature-based methods. Those features on geometry structures can be captured in 2D and 3D data and are not affected by dynamic elements like different illumination.


\setlength{\textfloatsep}{8pt}

\begin{figure}
\centering
\subfigure[Spinning LiDAR]{
\label{fig:device:a} 
\includegraphics[width=1.53in]{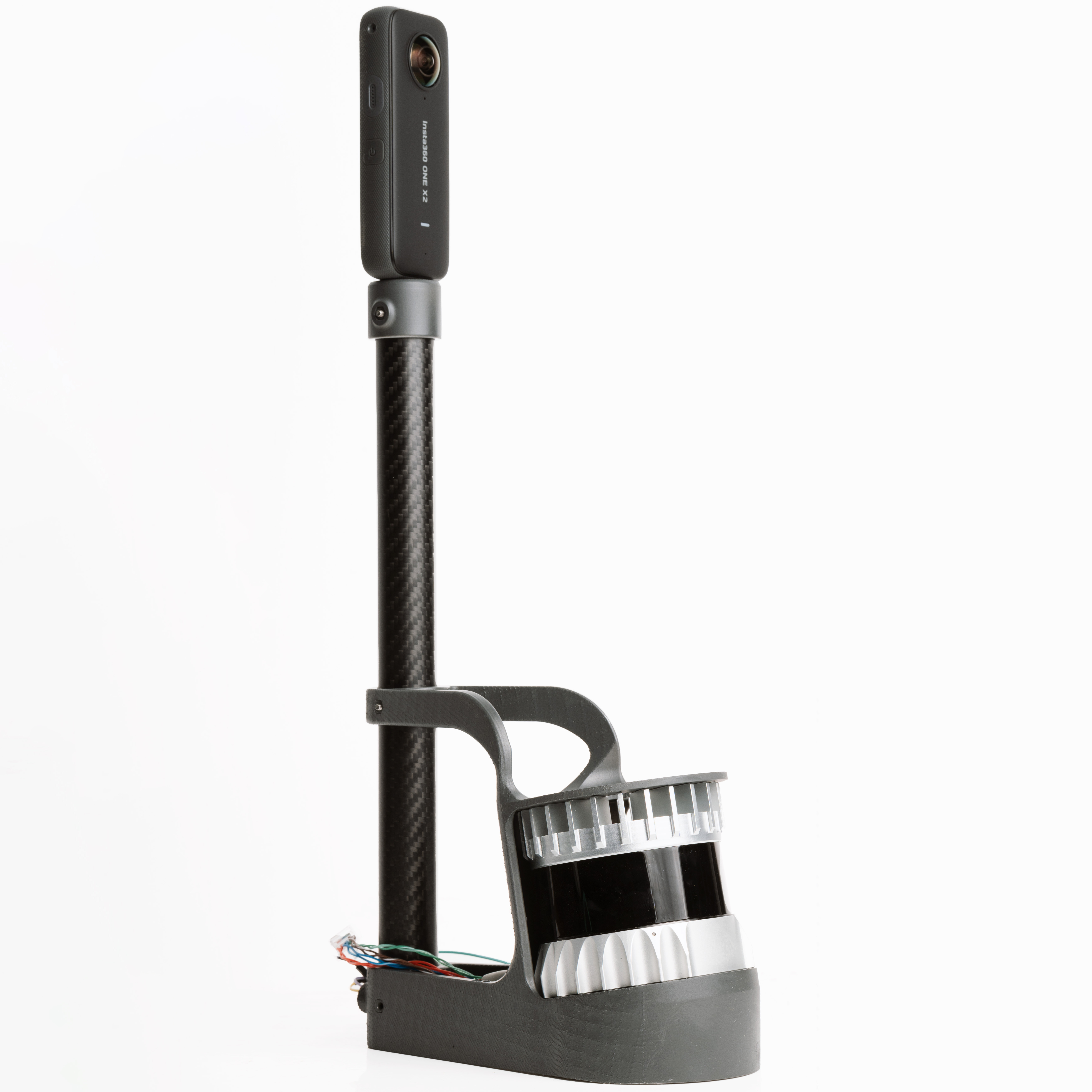}}
\subfigure[Solid state LiDAR]{
\label{fig:device:b}
\includegraphics[width=1.53in]{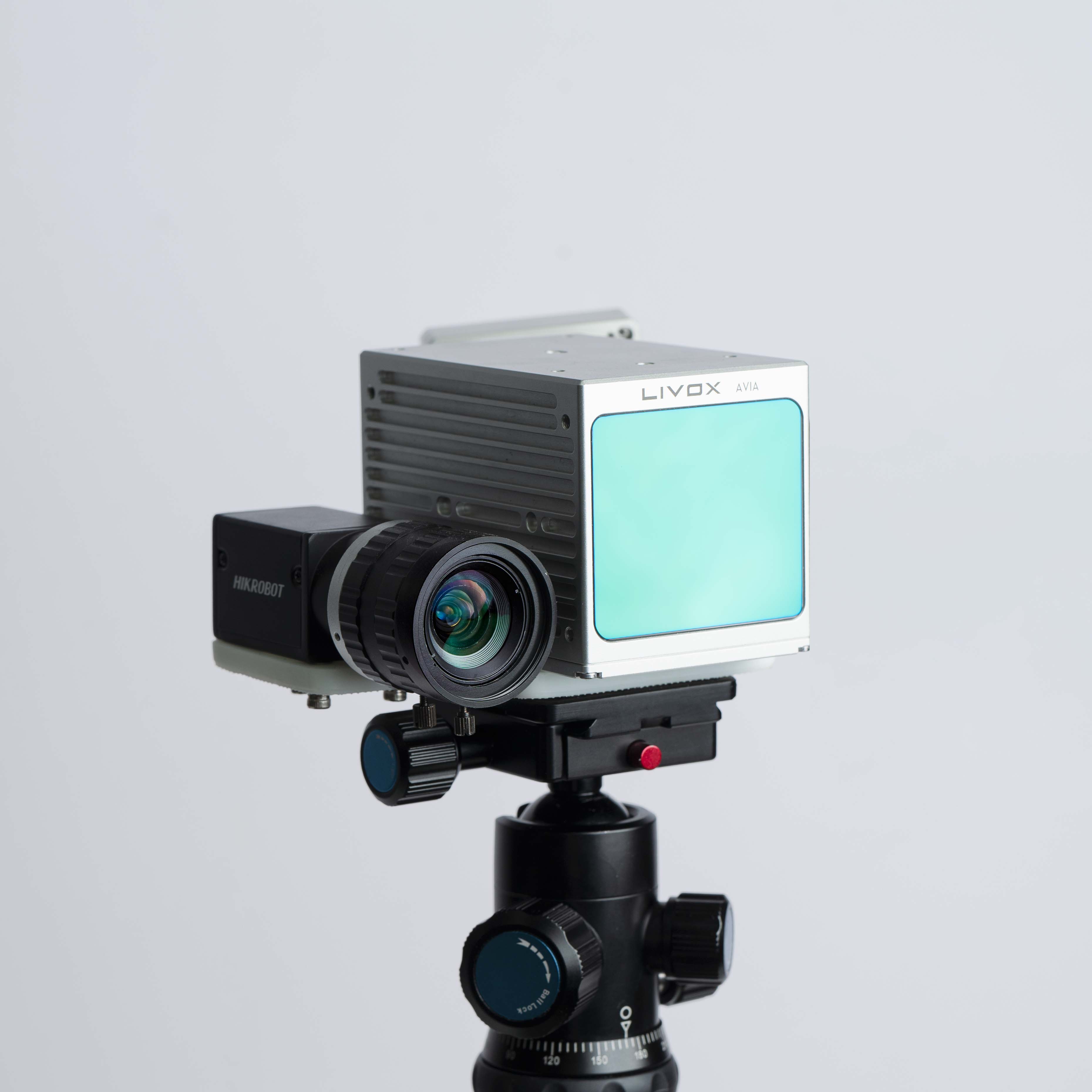}}
\subfigure[The input image]{
\label{fig:first_res:a} 
\includegraphics[width=1.53in]{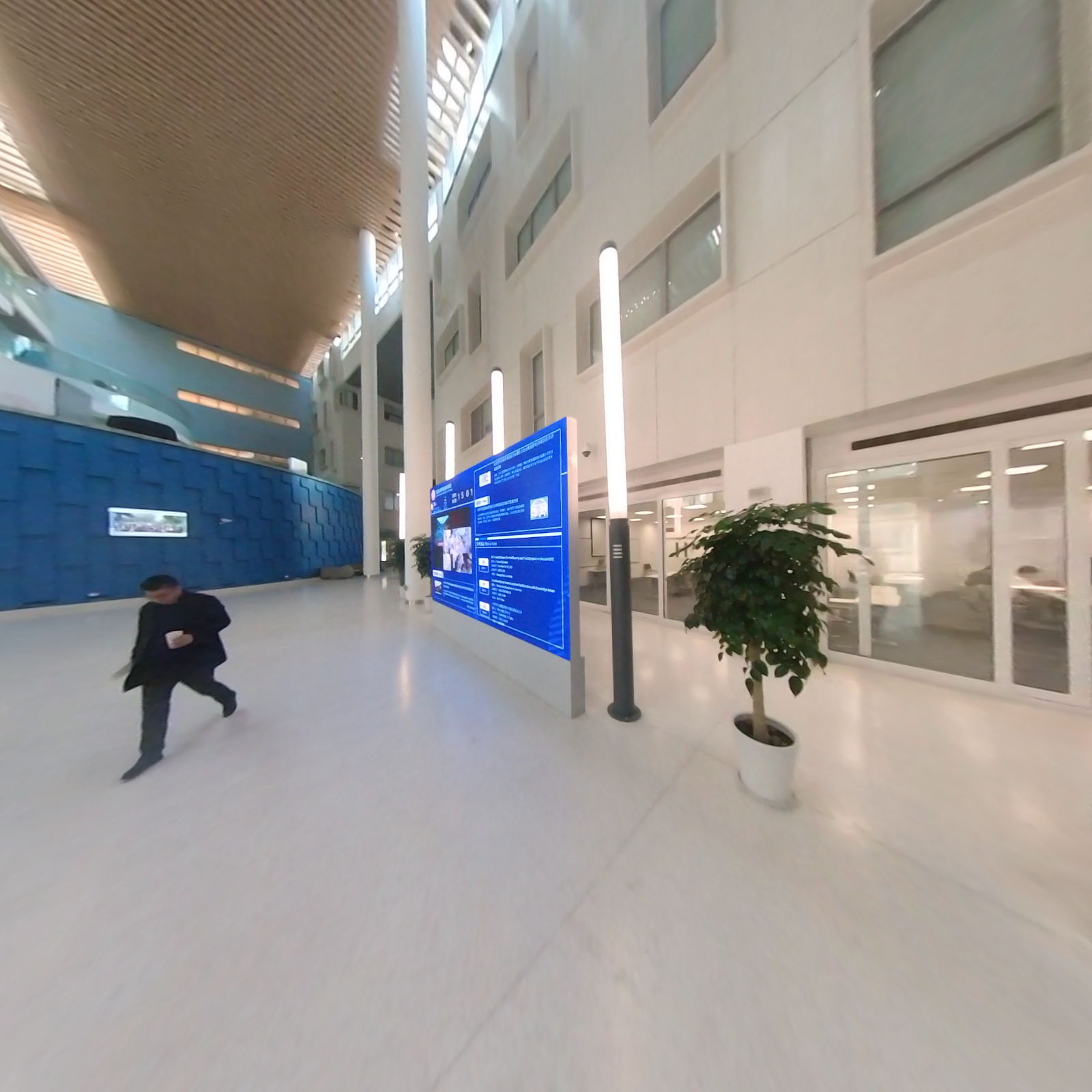}}
\subfigure[ROI of Point cloud]{
\label{fig:first_res:b}
\includegraphics[width=1.53in]{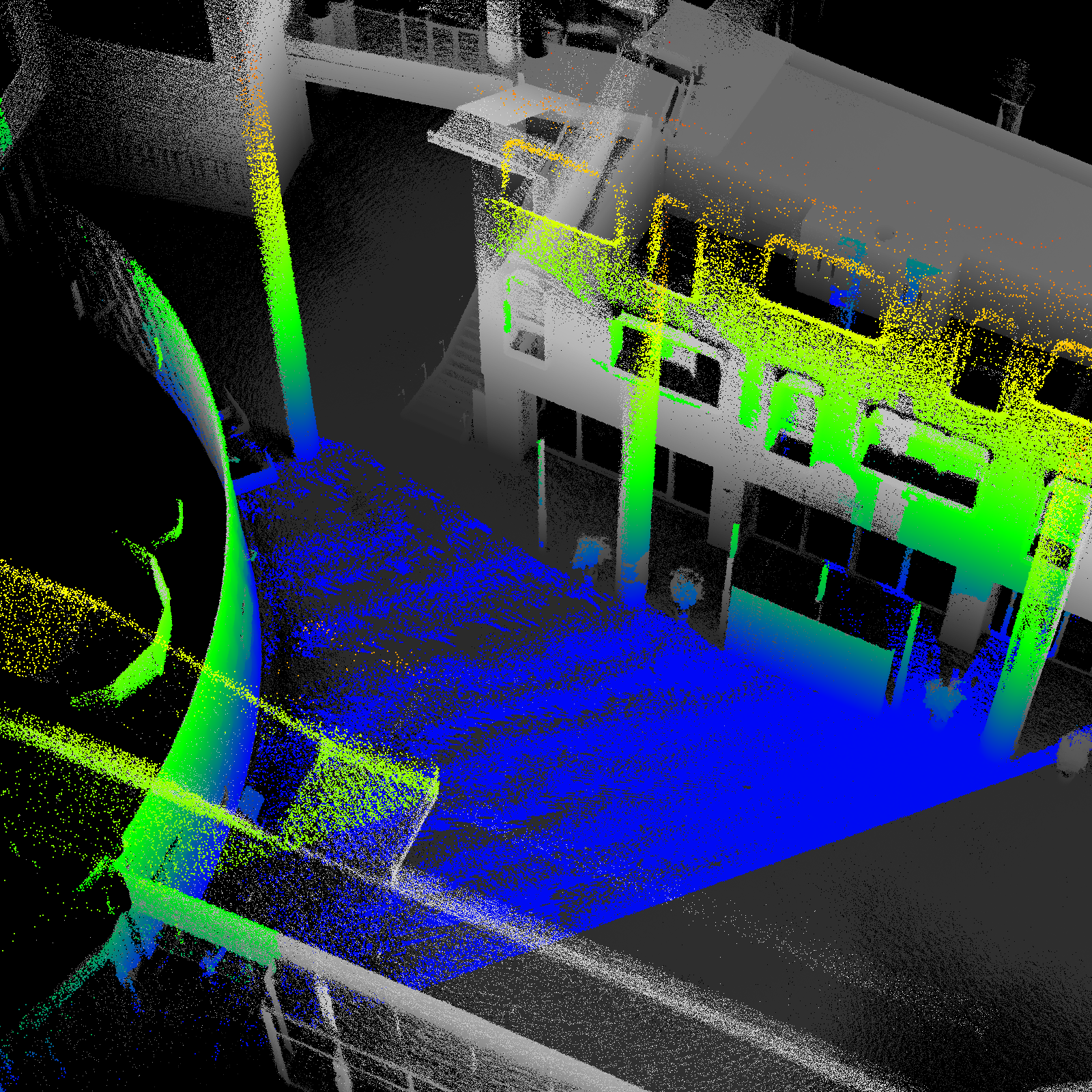}}
\caption{Our handheld devices for data collection: (a) The spinning LiDAR with rolling-shutter panoramic consumer camera with two fish-eye sensors; 
(b) Solid state LiDAR with global-shutter camera with a wide-angle low-distortion lens inspired by \cite{lin2021r3live}. Below the examples of input data: (c) is the input camera image; 
(d) is the input point cloud. The height-encoded colored point cloud is region of interest (ROI) we filtered, and the grey one is the initial input generated by LIO-SAM \cite{shan2020lio}.
}
\label{fig:device}
\end{figure}

Furthermore, because of the richness of edge features in most scenes, extraction and matching of geometric edges makes it possible to solve the 2D-3D registration problem. Nevertheless, direct 2D-3D geometric line correspondence matching does also easily lead to some registration error.

To overcome the above problems and challenges, and thanks to existing methods like \cite{lin2021r3live}, which can provide not only a good LiDAR mapping result but also a series of coarse camera poses, we propose a new framework for camera localization augmentation with the LiDAR mapping result for the purpose of point cloud colorization. Because the degradation situations are to be avoided as much as possible, in addition to actively handling these cases, some pre-processing can be performed on the input point cloud and images, such as removing blurry images and removing dynamic objects \cite{schauer2018peopleremover} from the input point cloud. In addition, depending on coarse camera poses and intrinsic parameters, 2D-3D line matching can be accelerated by selecting a suitable pending point cloud (ROI).

We have done experiments to test our algorithm, which illustrate that our algorithm has a significant improvement in point cloud colorization. It works with different kinds of LiDAR, including spinning and solid state LiDAR and a camera (see Fig. \ref{fig:device:a} and \ref{fig:device:b}). We test the accuracy of our algorithm and other camera re-localization methods on the same dataset, while using LiDAR-Inertial-Visual Odometry and Visual-Inertial Odometry (VIO) algorithms as baselines. 


The main contributions of our work are:
\begin{itemize}
\item We propose and implement a novel approach for monocular camera pose improvement in prior LiDAR maps, which is capable of handling reconstruction tasks in large-scale scenes.
\item Multi-layer voxelization is adopted in the framework. At each step of the LiDAR processing flow, including dynamic object removal, ROI filtering, and 3D line extraction, the multi-layer voxel maps data structure is used for input and output as a whole instead of the initial point cloud, which simplifies and speeds up the algorithm.
\item We carefully study and analyse numerous degraded optimization situations and failure cases for re-localization.
\item Various experiments have been conducted to test our method, which show the robustness and accuracy of our system.
\end{itemize}

\section{Related Work}


In addition to the PNP-based methods and SFM-based methods mentioned above, traditional Visual-Inertial Odometry (VIO) methods \cite{qin2018vins} and RANSAC-based or Branch-and-bound methods \cite{fischler1981random, campbell2018globally} can provide the camera poses in the global map. However, these methods rely heavily on the point correspondences to handle 2D-3D matching problem, and are also causing the problem of missing scale.

The 2D and 3D lines detection from images and the point cloud, respectively, and the according matching method is an intuitive approach to solve the visual based localization problem. Unlike using features, the edges or planes have higher-order semantic information. \cite{yu2020monocular} presented one of these kinds of methods based on VIO and it optimizes the camera poses obtained from the VIO algorithm.


Learning-based methods \cite{chang2021hypermap, feng20192d3d, cattaneo2020cmrnet++} are also designed to obtain the better pose estimates, directly registering the images to the LiDAR information. But their results are not stable on the camera localization tasks.

Our work is inspired by 2D-3D matching for camera localization \cite{yu2020monocular}. However, it considers all 3D edges extracted in the input point cloud and uses a learning-based method to extract 2D lines in the image. Thus, the applicability and robustness of the algorithm are relatively weak. We improve upon this by adding a ROI filtering operation and classifying the extracted 3D lines before 2D-3D matching. In \cite{yuan2021pixel, liu2021fast} the target-less calibration method was identified to provide comparable results with the state of the art (SOTA) traditional target-based calibration method, and in this paper we adopt the idea of line detection and matching in the target-less calibration method and compare with other re-localization or mapping methods.


\begin{figure*}
\centering
\includegraphics[width=0.95\linewidth]{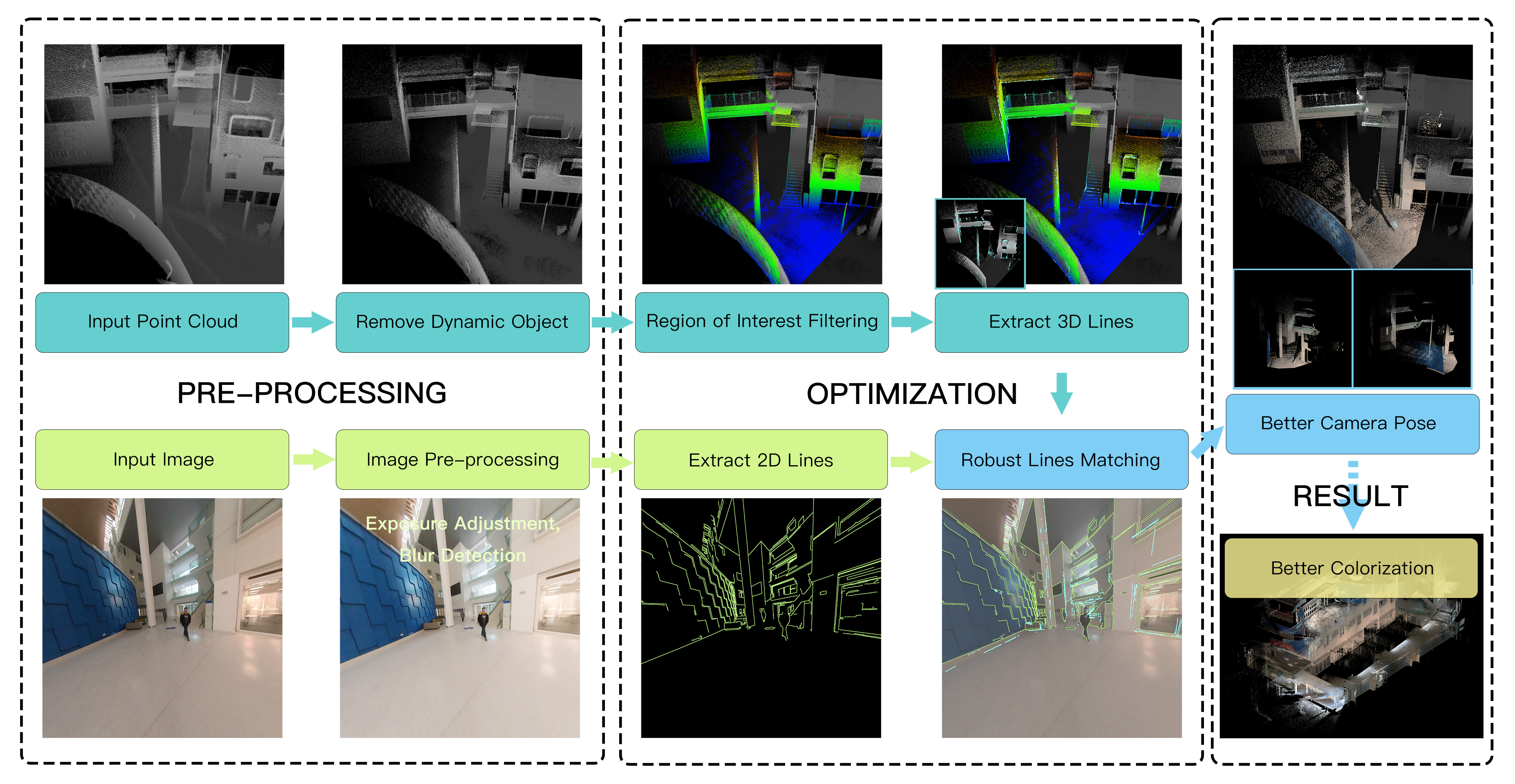}
\caption{The pipeline of the system. The light-green lines (Step \textit{Extract 2D lines} and \textit{Robust Lines Matching}) denote the edges detected by Canny algorithm from the input image, and the cyan lines (Step \textit{Extract 3D lines} and \textit{Robust Lines Matching}) are the depth-continuous edges \cite{yuan2021pixel} detected from the input point cloud of ROI.}
\label{fig:pipeline}
\vspace{-0.5cm}
\end{figure*}

\section{Overview of the System}

The overview our framework is shown in Fig. \ref{fig:pipeline}, which contains two main processing flows for image and point cloud input. Before these two flows, the LiDAR odometry and mapping algorithm \cite{shan2020lio, lin2021r3live} is implemented to generate a global and correct-enough point cloud map. Here, the LiDAR-only bundle adjustment \cite{liu2021balm} is also an optional pre-processing step to enhance the point cloud results. If we use a LIO-based method, we can obtain correct-enough LiDAR poses, which help to get a smooth and almost-accurate trajectory of the motion with the high frequency IMU data. Relying on the rough extrinsic parameter and time synchronization, the coarse camera poses are then calculated.

For the image input, we will determine whether it is blurry or not. The image will be thrown away if it is, because existing image deblur algorithms do not work well. Furthermore, some digital image processing methods like histogram equalization is used first, in order to set a unified threshold to judge the blur situations. It is worth noting that in most cases, when we collect data, exposure is usually fixed. Therefore, in order to ensure the continuity of the color of the point cloud, we only use the original images for the colorization test without some special pre-processing. After that, we detect 2D edges with the Canny algorithm.

For the global point cloud input, we will remove dynamic objects first, which is an upfront work in the pipeline if we want to handle a series of camera poses with corresponding images. Then we select the point cloud ROI based on the camera view. Only the depth-continuous edges will be extracted by intersecting planes detection, because depth-discontinuous edges in 3D point clouds are often not reliable for matching 2D-3D lines.

With 2D and 3D lines extracted by two processing flows, a matching optimization algorithm is used to obtain the optimal camera pose parameters of the 2D image in the 3D point cloud. We also deal with some degraded optimization problems in this step. The optical image textured point cloud of ROI can be rendered using the camera pose parameters.

If this framework is employed in a small-scale local scene, the removing dynamic objects and ROI filtering parts will be skipped, which will speed up the pipeline of 2D-3D correspondences matching and providing a near real-time performance. 

\section{Methodology}

\subsection{Multi-layer Voxelization}

The main idea is inspired by the concept of adaptive voxelization \cite{liu2021balm, liu2021fast} and OctoMap \cite{hornung2013octomap}. In our approach, the previous work of dynamic object removal \cite{schauer2018peopleremover}, ROI filtering \cite{katz2015visibility}, and 3D line extraction \cite{liu2021fast} work with a voxel-like data structure, thus we implement a multi-layer voxelization process for versatility and simplicity of data transmission.

Our voxel map is constructed by an Octotree data structure. The roots here are the results of traditional voxelization for 3D point cloud, which means we voxelize the point cloud by a default and slightly larger size (e.g, $2m$) first.
Then all roots/voxels will be divided into eight child nodes/ small voxels (child nodes are half the length of the parent node), until the voxel will reach the proper minimal size (e.g, $62.5mm$). Only all leaves store the corresponding small point cloud data in order to save storage.
The voxel map helps to accelerate the preliminary selection of the point cloud, and also simplifies the initial point cloud data.

\subsection{Dynamic Object Removal}



We adopt the ray-tracing based method presented in \cite{schauer2018peopleremover} to filter the dynamic objects. But unlike traditional ray-tracing based methods, the algorithm will compare simultaneously various scans and apply some additional heuristics.

The main idea of the ray-tracing based method for filtering dynamic objects is to construct a global occupancy grid stored as a voxel data structure for point clouds and to determine whether each grid/voxel will be hit or seen through. Then the voxel will be viewed as a dynamic one, if it is first hit by a LiDAR scan, and then is seen through by the optical path of other LiDAR scans. The point cloud in this voxel will be filtered. Most of the previous algorithms \cite{asvadi20163d} adopted the idea with directly storing the numbers of hits or see-throughs into each voxel, thus they contained some false positive cases, the static voxels are incorrectly marked as dynamic ones, caused by LiDAR spots with approaching angles in the same scan (self-intersection cases). To deal with this situation, each voxel stores a set of LiDAR scan identifiers. For each voxel, we will traverse all voxels on an optical path from the origin of the LiDAR to it (ray-tracing), if one of these voxels contains the same scan identifier, the voxel will not be considered to be a dynamic one.

In addition to the above-mentioned indirect processing of voxel occupancy situations, there are still a lot of false positive cases, which mainly include two kinds of situations due to heterogeneous sampling of surfaces for LiDAR scanning. One is when the incident angle is relatively large, the LiDAR scan will pass through the voxels where the LiDAR spots with a small incident angle is located; The second one is in the occlusion case, where the LiDAR scan at the edge of the foreground object will be easily passed through by other scans. Thus, the setting of safe maximum search distances \cite{schauer2018peopleremover} for each LiDAR spot can avoid these false positives. Voxels below the maximum distance are not filtered. 


In this part, we build the voxel map at first and set a suitable minimal size of voxel (e.g, $62.5mm$). Then dynamic object removal with the highest-layer of voxel map will prune branches in this map.

\subsection{Region of Interest Filtering for point cloud}


The main concept of this part is to find the visible part of the point cloud from the viewpoint, which is the application of the Generalized Hidden Point Removal (GHPR) operator \cite{katz2015visibility}. It mainly contains two steps: point transformation and convex hull construction.  

A naive mirror kernel will be applied to speed up the processing for visibility checking of the point cloud, which can be written as 
\begin{equation}
    f(d) = \phi - d,\ \phi \geq max_{p_i \in P} (\Vert p_i\Vert).
\end{equation}
Here, the point $p_i$ is in the point set $P \subset \mathbb{R}^ 3$ sampled from a continuous surface and $d = ||p_i||$. The equation indicates flipping around a spherical mirror with radius $R = \frac{1}{2} \phi$, and $C$ denotes the center of the sphere. Then a radius transformation \cite{katz2007direct} can be defined as 
\begin{equation}
    F(p_i) = p_i + 2(R - \Vert p_i\Vert) \frac{p_i}{\Vert p_i\Vert}.
\end{equation}
The new point set $\hat{P} \subset \mathbb{R}^ 3$ will be obtained by applying this radius transformation. 
The points reside on the convex hull of $\hat{P} \bigcup C$ and determine which points are reachable or not, representing  $\hat{P}^* \subset \hat{P}$. Then we find the relevant reachable point subset $P^* \subset P$.  Depending on the camera intrinsic parameters, the points selected finally are the visible points. 

In order to simplify and speed up the work flow, we select some root nodes in the voxel map with a preset maximum search distance (e.g, $20m$), that is, part of the point cloud to be processed. Apart from pre-selection, we do not consider every point in the pending point cloud to filter our ROI, but consider nodes/voxels in higher layers in the voxel map as a whole. 
Without loss of generality, we take all leaf nodes from selected root nodes into account. After that, we down-sample the point cloud by treating each leaf node/voxel as a point (i.e. the center of the voxel), which accelerates the process of computing the convex hull. A few points that would have been filtered out will also be retained, but these points should be retained for later processing due to the inaccurate camera poses. 
It is for this reason that some operations about point cloud expansion need to be implemented in practice. First we properly widen the field of view (FOV) of the viewpoint before filtering.
Then each visible voxel is also traversed and its adjacent voxels are marked as visible, further expanding the ROI filtering results. Fig. \ref{fig:first_res:b} and \ref{fig:first_res:a} show the point cloud of ROI from the camera viewpoint and the corresponding image.

Because occlusion detection of 3D line features is difficult, the 3D line maps can not directly select ROI for 3D lines easily. Some methods \cite{yu2020monocular} collect all visible 3D lines without discarding occluded ones, and rely only on the camera intrinsic parameters and poses information. Consequently, there will be many occluded lines when doing with a large-scale point cloud. 

\begin{figure}
\centering
\subfigure[Image with 2D lines]{
\label{fig:occlusion:a} 
\includegraphics[width=1.04in]{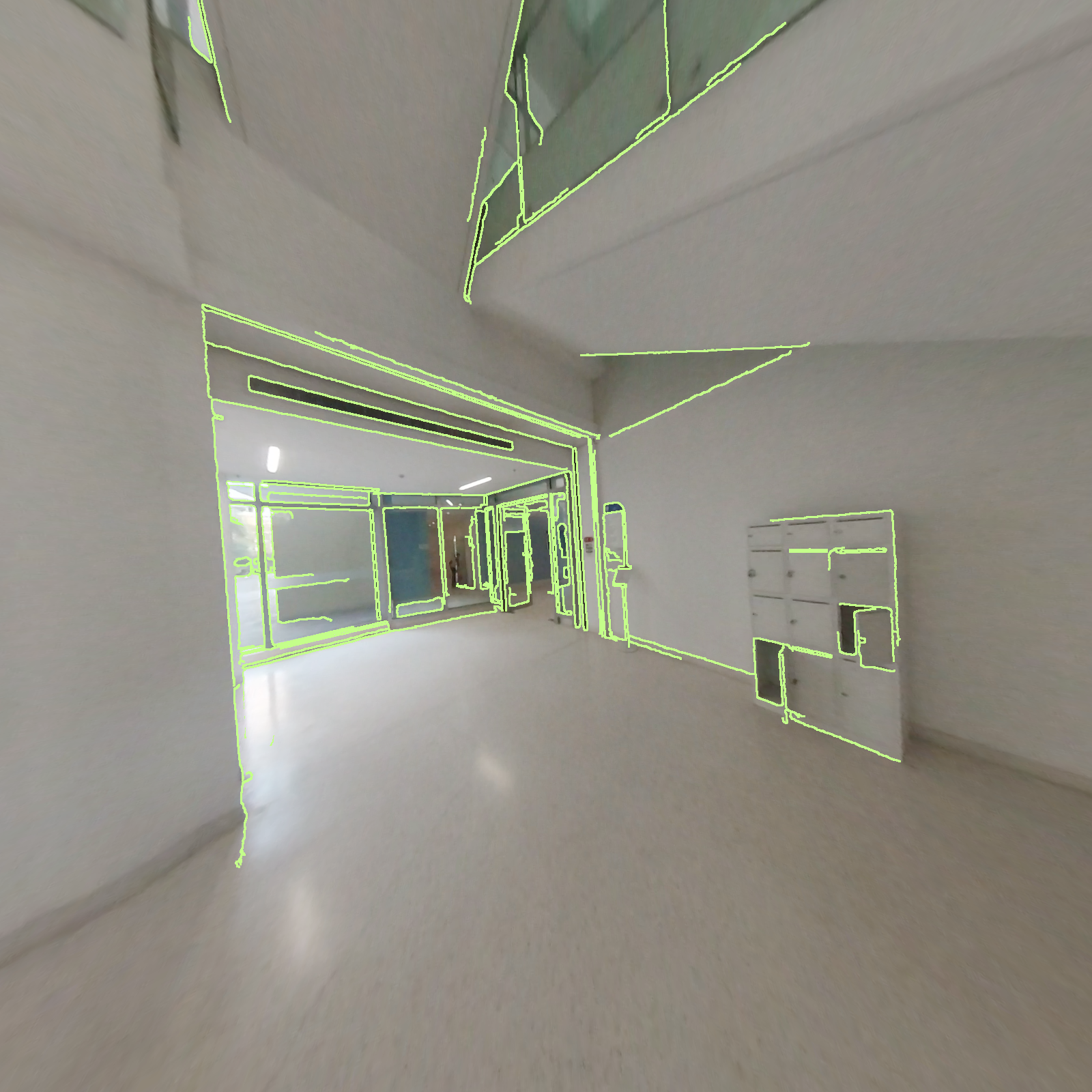}}
\subfigure[After ROI filtering]{
\label{fig:occlusion:b}
\includegraphics[width=1.04in]{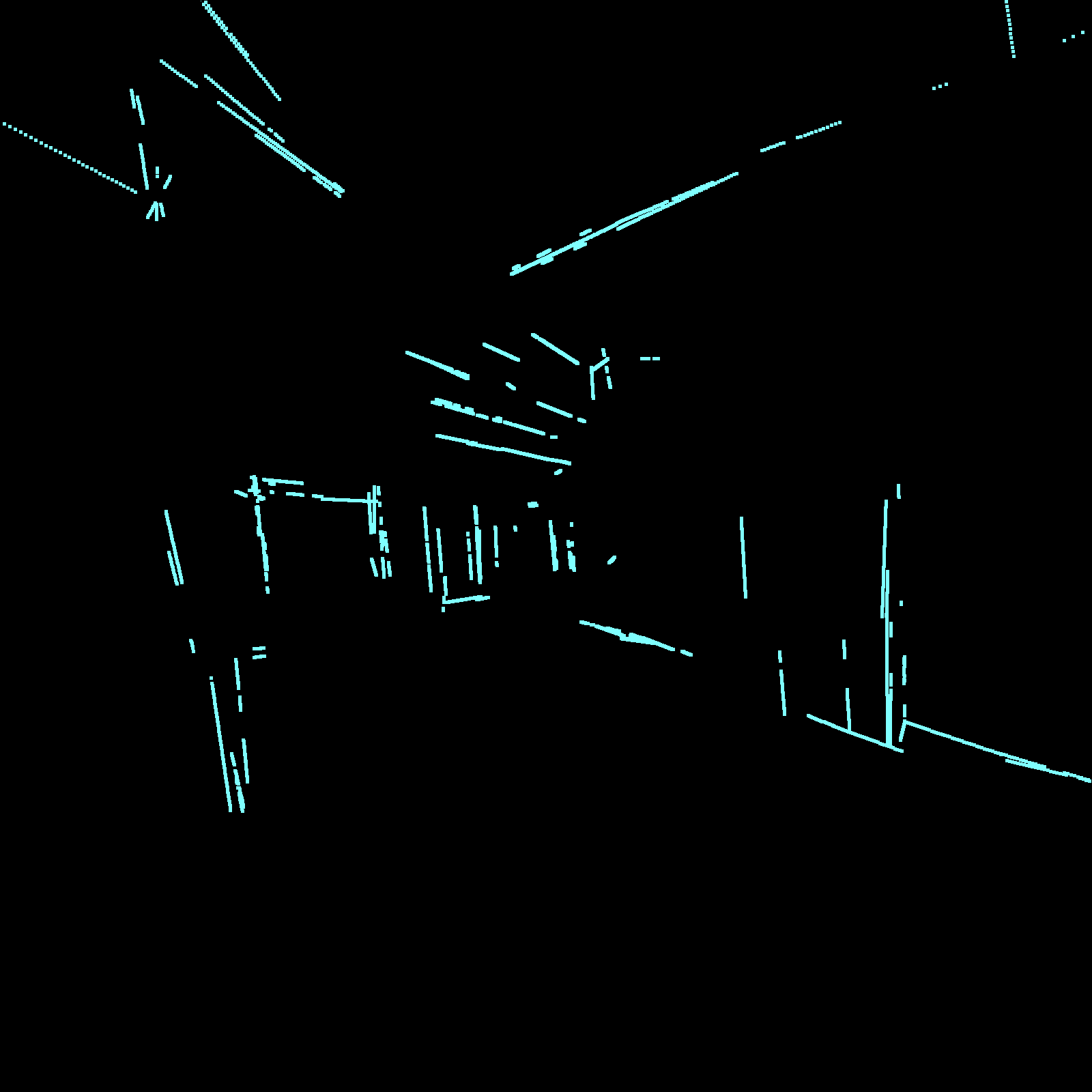}}
\subfigure[Before ROI filtering]{
\label{fig:occlusion:c}
\includegraphics[width=1.04in]{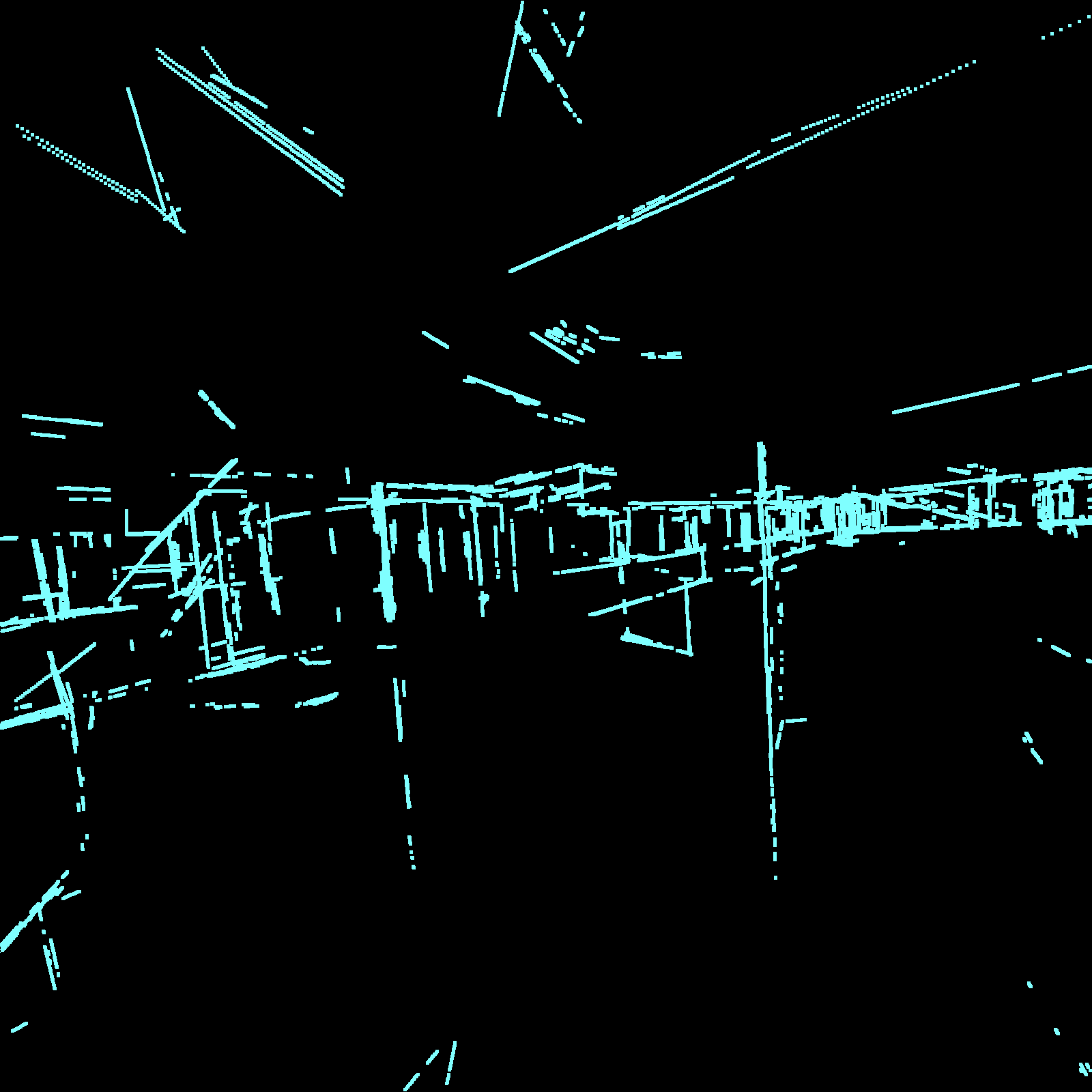}}
\subfigure[Overall situation of the scene]{
\label{fig:occlusion:d}
\includegraphics[width=3.26in]{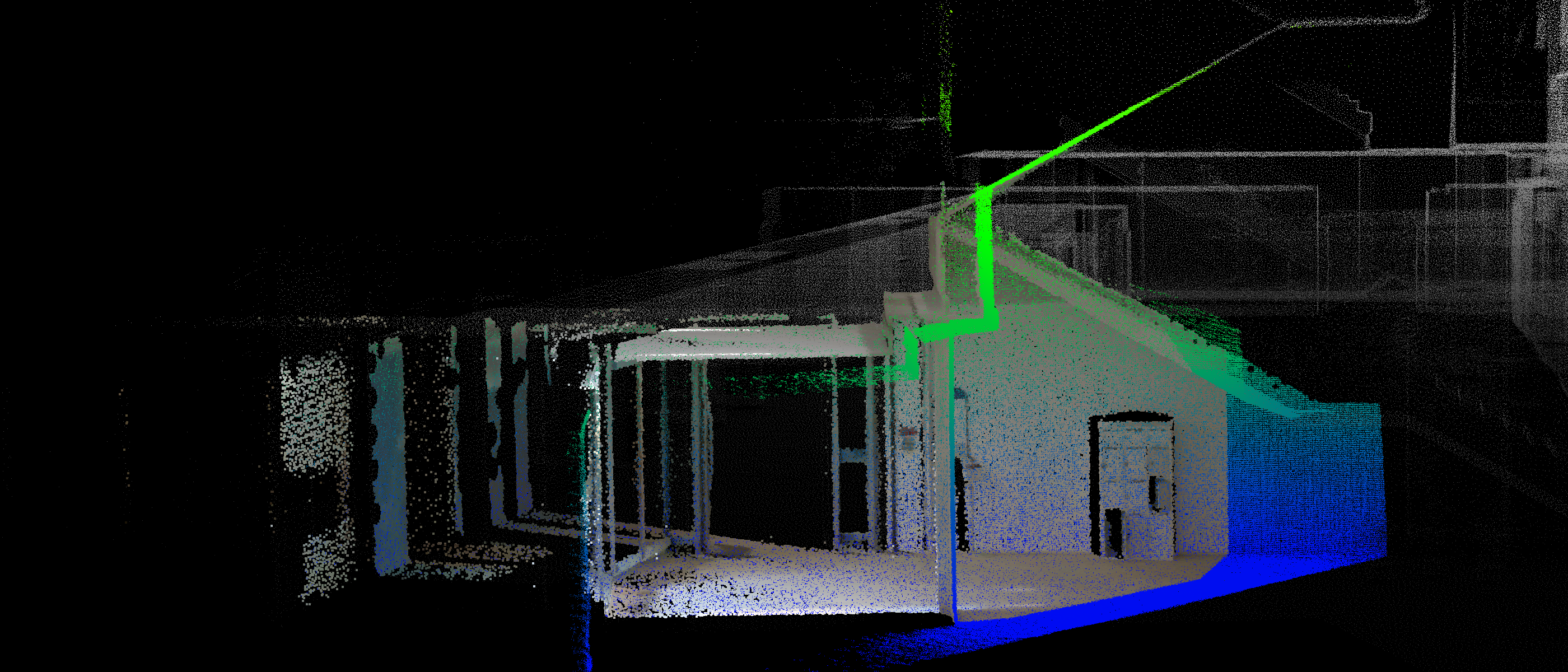}}
\caption{The comparison about occlusion detection: (a) is the input image with 2D light-green lines; 
(b) is the 3D line extraction result after ROI filtering;
(c) are the 3D lines extracted directly from the whole point cloud;
(d) is the ROI selection result and final colorization result. The grey-scale point cloud is the input point cloud, which is down-sampled in order to illustrate the cases clearly. The height-encoded (green-to-blue) point cloud is the ROI, and the RGB-colored one is the final result.}
\label{fig:occlusion}
\end{figure}

This situation is indicated by Fig. \ref{fig:occlusion}. In this case, we can find there exists a corridor behind the visible wall (Fig. \ref{fig:occlusion:d}), so the 3D lines extracted in the initial point cloud are shown in Fig. \ref{fig:occlusion:c}, which greatly increases the difficulty of 2D-3D correspondence matching,  even though all depth-discontinuous edges are already excluded. Based on this consideration, we will filter the point cloud of ROI according to each image pose and then extract the 3D lines, rather than directly extracting all 3D lines in the beginning.

\subsection{3D Line Detection}
\begin{figure}
\centering

\includegraphics[width=2.42in]{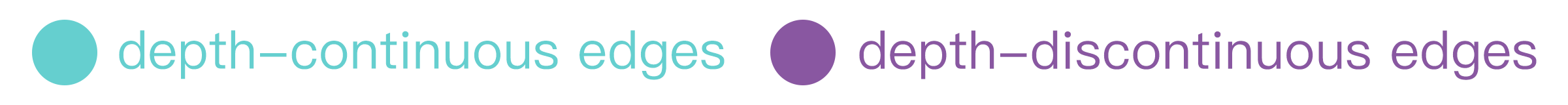}

\subfigure[3D RGB-colored point cloud]{
\label{fig:edge:a} 
\includegraphics[width=1.92in]{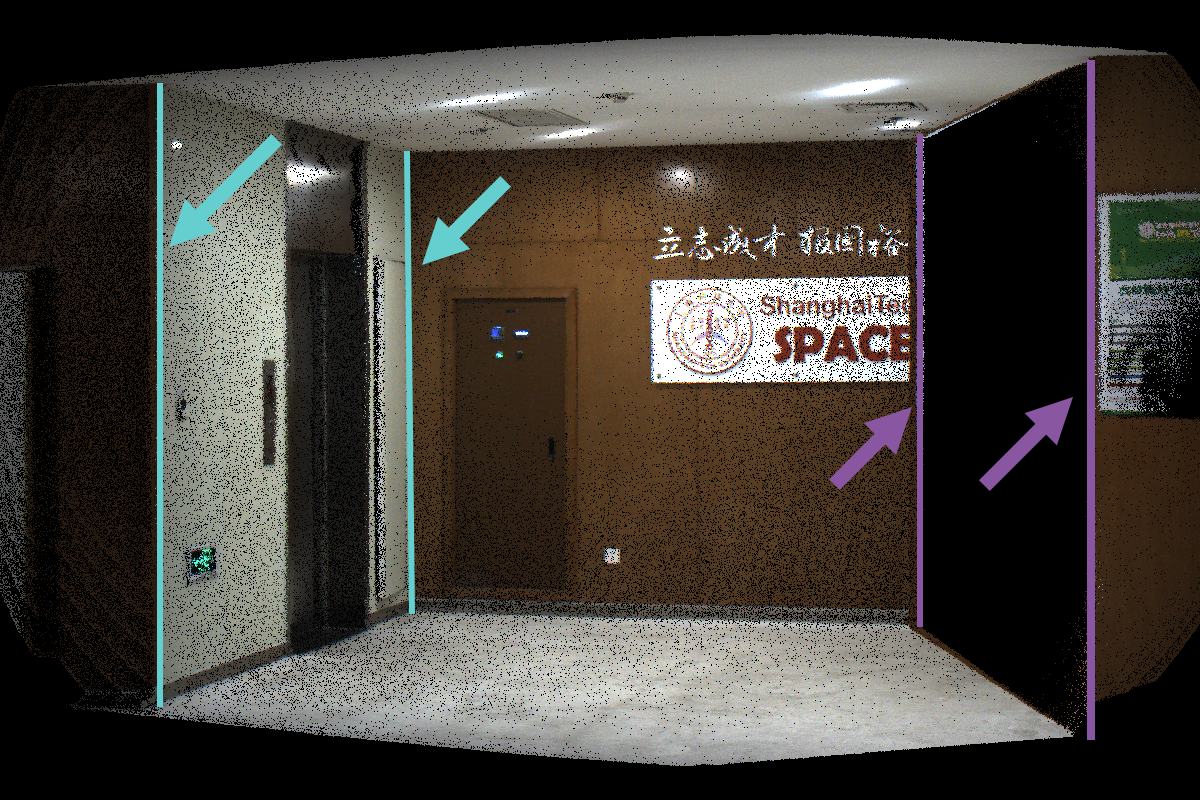}}
\subfigure[2D corresponding image]{
\label{fig:edge:b}
\includegraphics[width=1.28in]{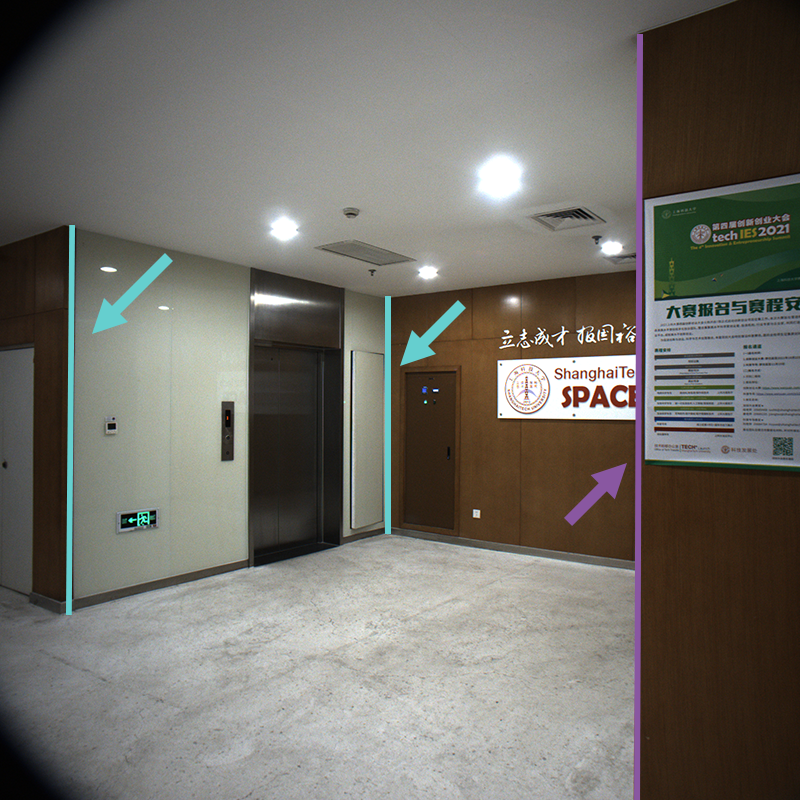}}

\caption{Depth-continuous and depth-discontinuous edges.}
\label{fig:edge}
\end{figure}

In general, there exist two types of feature edges: depth-continuous and depth-discontinuous edges, as displayed in Fig. \ref{fig:edge}.
The depth-discontinuous edges are prone to the foreground inflation and bleeding points, resulting in the zero- or multi-value projection problems. It is thus intuitive that the depth-discontinuous edges are not reliable for matching with the image information. Thus, to acquire depth-continuous ones, we will detect two intersecting planes beforehand.

Similar to the method presented in \cite{liu2021fast}, we traverse all root nodes in the range of ROI. With regard to each point $q_i \in Q \ (i = 1, ..., m)$ contained in a node, the covariance matrix $\textbf{K}$ of a $3$-dimension vector $q_i$ is
\begin{equation}
    \textbf{K} = \text{var}(Q) = \frac{1}{m} \sum^m_{i=1} (q_i - \textbf{E}(q_i)) (q_i - \textbf{E}(q_i)) ^T,
\end{equation}
where $\textbf{E}(q_i) = \frac{1}{m} \sum^m_{i=1} q_i$. Three eigenvectors of $\textbf{K}$ represent the main orthogonal directions of the point cloud and the corresponding three eigenvalues are proportional to dispersion in these directions. Therefore, if the smallest eigenvalue is close to $0$, the node/voxel will be considered as a planar one. In practice, calculating and comparing the ratio eigenvalues of a covariance matrix on these points is a more reasonable way to determine the voxel situations.

If a node/voxel does not contain a plane, its child nodes will be searched until the child is planar or all leaf nodes are traversed. The method is better than directly extracting planes in default size voxels by fitting with RANSAC \cite{yuan2021pixel}, because RANSAC can lead to wrong extractions. For each planar node, we scan all adjacent voxels. If the adjacent voxel is also planar and the angle between the normal vectors of two planes is within a range, the intersection edge can be constructed, which is a depth-continuous edge.

\subsection{Line Matching}


First we sample several global frame points $x_i \in X \subset \mathbb{R}^ 3 $ for each 3D edge we extracted before. With the coarse camera pose $\textbf{T}_c \in SE(3)$, the 3D-2D projection can be written as

\begin{equation}
    \hat{x}_i = \pi \biggl( \textbf{T}_c \begin{bmatrix} x_i \\ 1 \end{bmatrix} \biggr) \in \mathbb{R}^ 2,
\end{equation}
where $\pi (\cdot)$ is the function which implements the camera model and applies known camera intrinsic parameters. Then we will find the nearest neighbor $y_i \in Y$ of $\hat{x}_i$ in the 2D line extracted in the input image. In addition, the normal vector $\text{n}$ of this edge is considered to remove some common error matches (e.g, non-parallel but close lines).

With the 2D and 3D lines extracted, we formulate the problem for 2D-3D correspondence as: 
\begin{equation}
    \textbf{T}_c^* = \arg \min_{\textbf{T}_c, y_i} \sum^n_{i = 1} \text{n}^T \biggl(\pi \biggl( \textbf{T}_c \begin{bmatrix} x_i + \epsilon^{L}_i \\ 1 \end{bmatrix}\biggr) - (y_i + \epsilon^{C}_i )\biggr),
\end{equation}
where $\epsilon^{L}_i$ and $\epsilon^{C}_i$ represent the noise models \cite{yuan2021pixel} for LiDAR and camera, respectively. By minimizing the re-projection distance of the 2D-3D correspondences \cite{yuan2021pixel} the more accurate camera pose is obtained.








\begin{figure}
\centering
\subfigure[Point cloud for ground truth]{
\label{fig:metric:a} 
\includegraphics[width=1.53in]{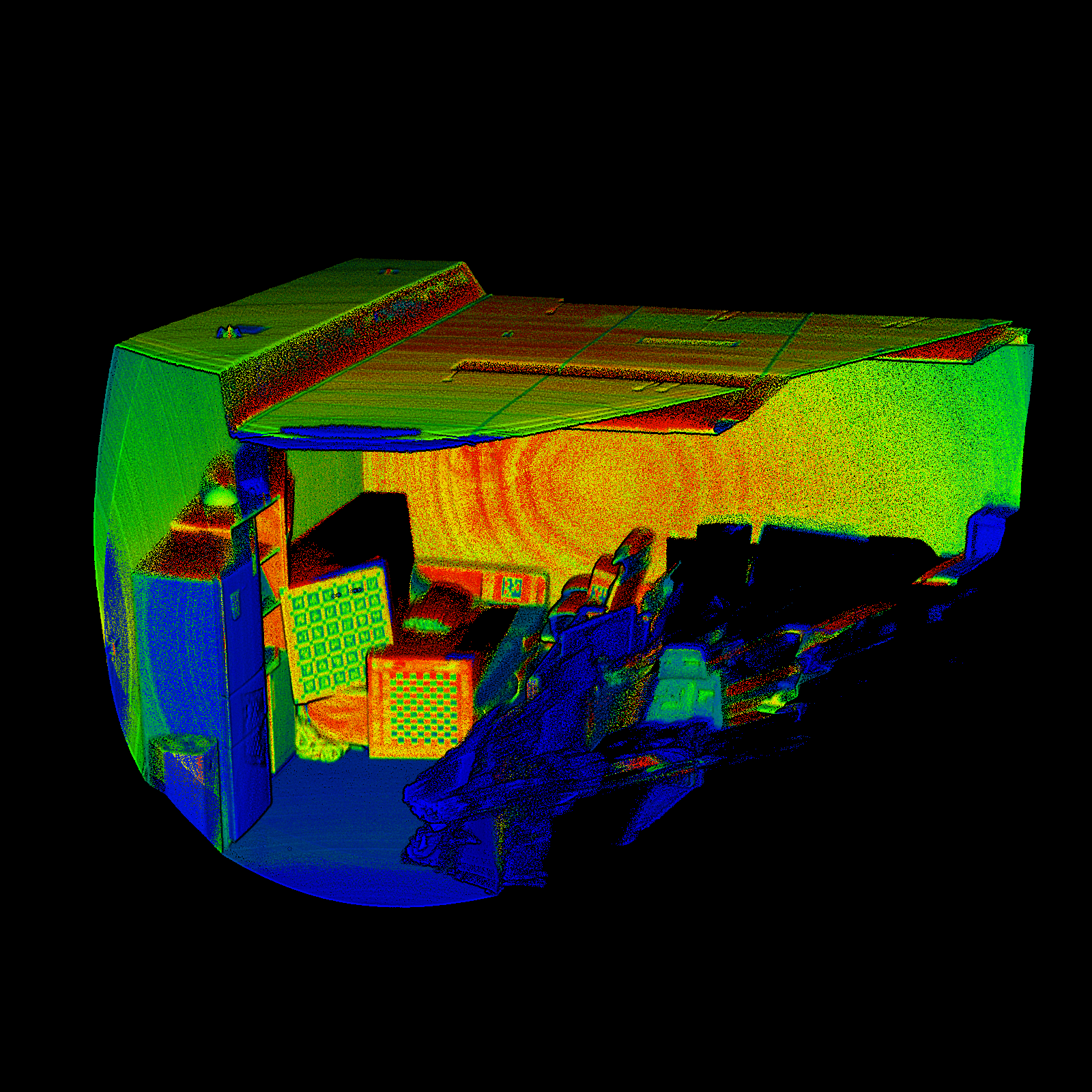}}
\subfigure[An input image]{
\label{fig:metric:b}
\includegraphics[width=1.53in]{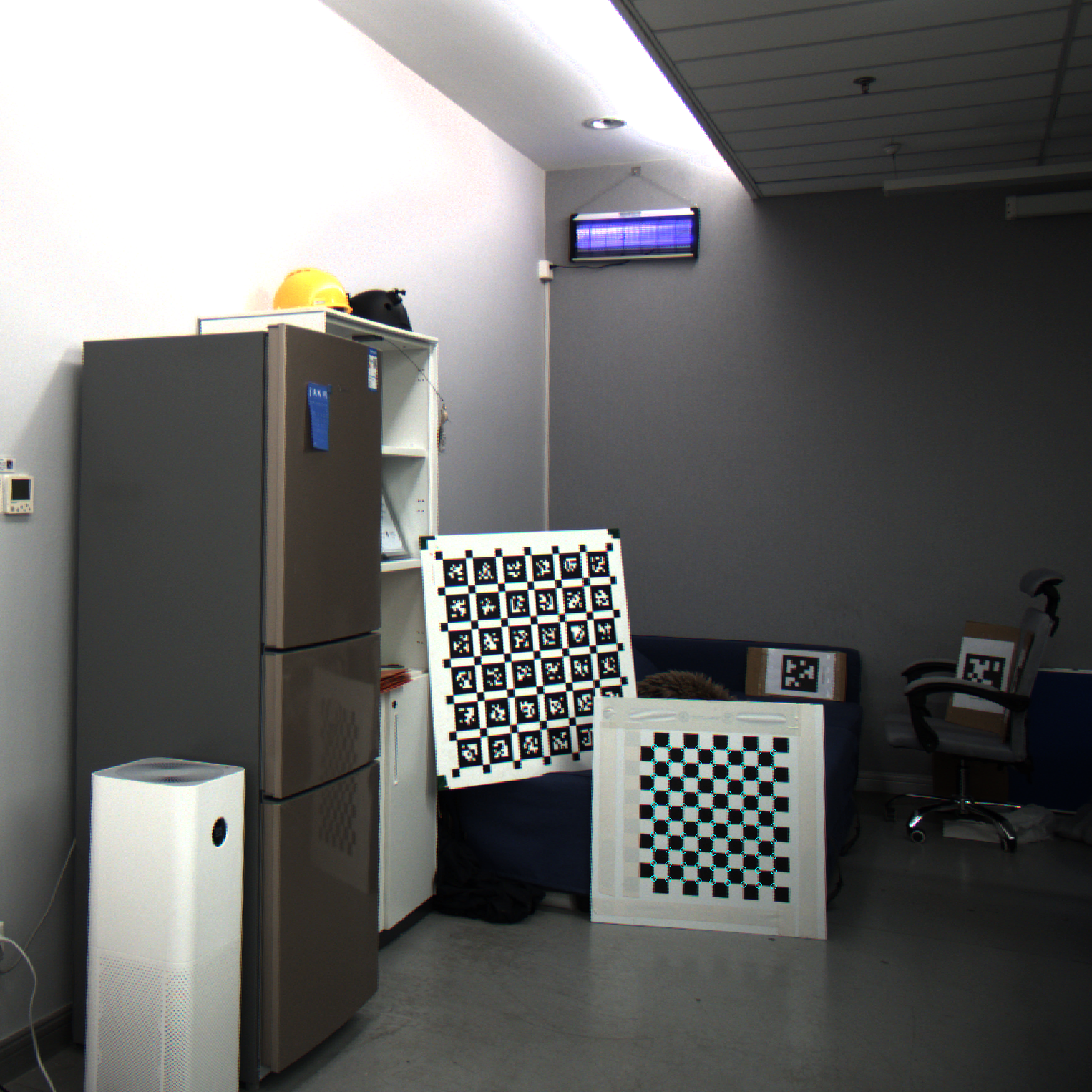}}
\\
\subfigure[Point cloud based on $\textbf{T}$]{
\label{fig:metric:c}
\includegraphics[width=1.53in]{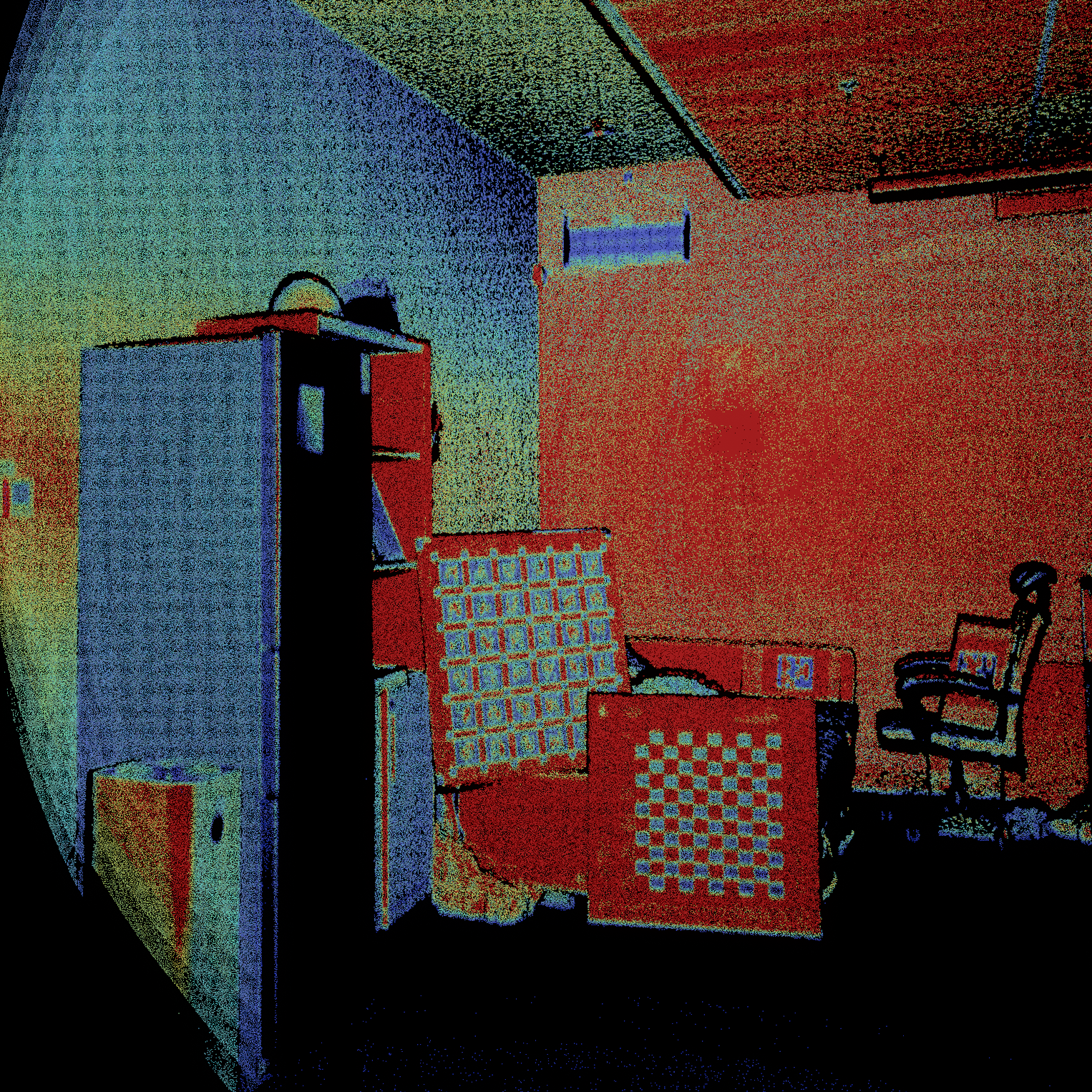}}
\subfigure[Residual visual display]{
\label{fig:metric:d}
\includegraphics[width=1.53in]{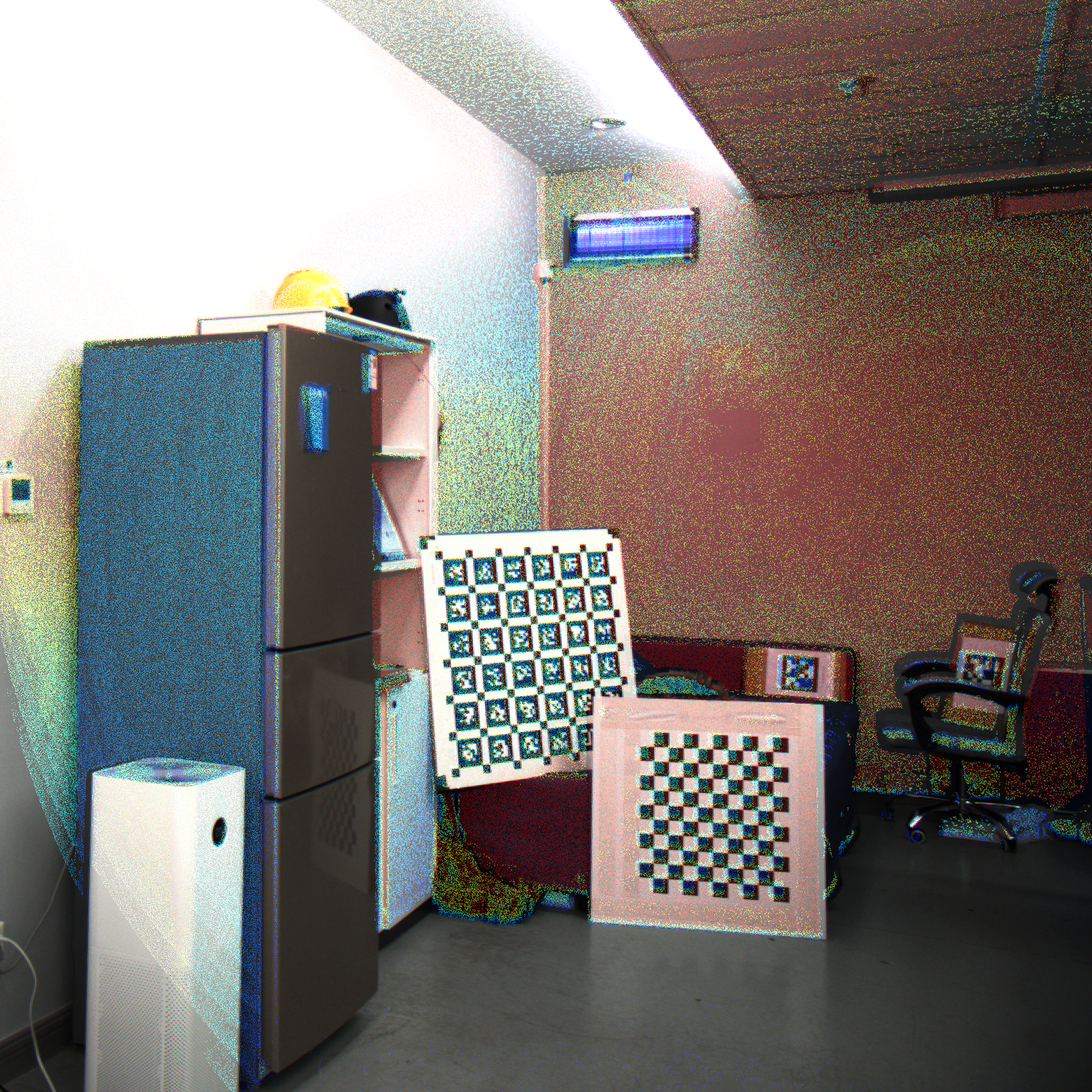}}
\\
\subfigure[Evaluation result]{
\label{fig:metric:e}
\includegraphics[width=3.3in]{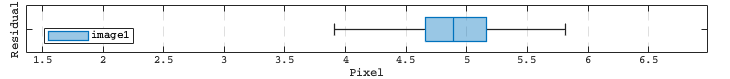}}

\caption{The evaluation of our experiment: (a) is the intensity-encoded colored point cloud; 
(b) is one example for an input image with checkerboard detection;
(c) The point cloud from the perspective of one camera pose $\textbf{T}$;
(d) is the visual display of residuals;
(e) is the $Residual(\textbf{T})$ of (d) calculated by Eq. \ref{eq:residual}.}
\label{fig:metric}
\end{figure}



\section{Evaluation and Results}

\subsection{Metric for Single Image Evaluation}



When using the solid state LiDAR, we first put the LiDAR in a stable pose for a period of time (e.g, $30s$), and collect an intensity-encoded colored point cloud (Fig. \ref{fig:metric:a}) as the ground truth, which contains one or more checkerboards. Then we start running the LIO or LIO\&VIO algorithm with this pose as the initial position, so as to obtain initial coarse camera poses $\textbf{T}_{init}$ and images. For the spinning LiDAR the ground truth is the result of LIO or LIO\&VIO algorithm. The dataset is collected by this process, which contains rooms, corridors, corners, etc., and the area is about 1,200 $m^2$. Finally, the optimized poses $\textbf{T}_{res}$ are obtained through our algorithm. For each pose $\textbf{T}$ whose image contains a checkerboard, with the camera model $\pi (\cdot)$, we have the residual of it:
\begin{equation}
    Residual(\textbf{T}) = \frac{1}{n} \sum_{i=1}^n \biggl \lVert p_i-\pi \biggl( \textbf{T} \begin{bmatrix} \hat{p}_i \\ 1 \end{bmatrix} \biggr) \biggr \rVert,
    \label{eq:residual}
\end{equation}
where $\hat{p}_i \in \hat{P} \ (i = 1, ..., n)$ denotes the corner point of the checkerboard we manually selected in the point cloud by its intensity information, and $p_i \in P \ (i = 1, ..., n)$ is the corner point of the checkerboard detected in the image. Fig. \ref{fig:metric} shows the evaluation for one case in a scene. For the scene placement, we placed the checkerboard in a position with many edges, especially the corners. 

\subsection{Augmentation of camera poses based on LIO Algorithm}






We run our method based on LIO-SAM \cite{shan2020lio}, and test other localization or re-localization algorithms. It is worth noting that some parts of the 2D-3D matching method based on \cite{yu2020monocular} were rewritten to fit with our data, but the performance is basically in line with the test in \cite{yu2020monocular}, especially compared to VINS-Mono \cite{qin2018vins}. In addition, similar with calculating relative poses errors (RPE) \cite{zhang2018tutorial}, the poses of 2D-3D matching and VINS-Mono are aligned with the smooth trajectory built by LiDAR and high-frequency IMU data.



\begin{figure}
\centering
\includegraphics[width=0.49\linewidth]{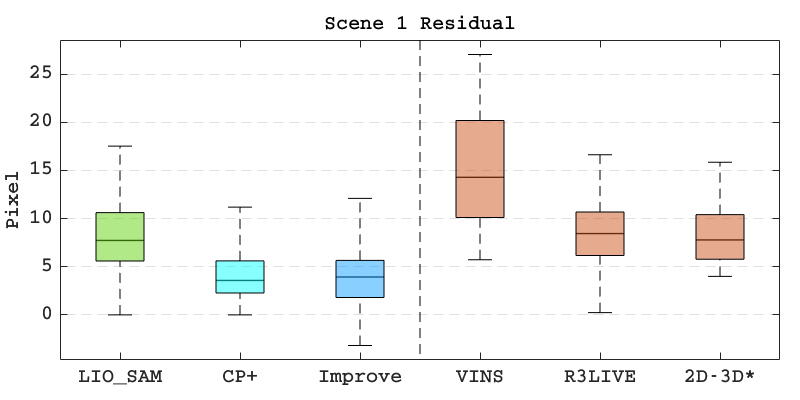}
\includegraphics[width=0.49\linewidth]{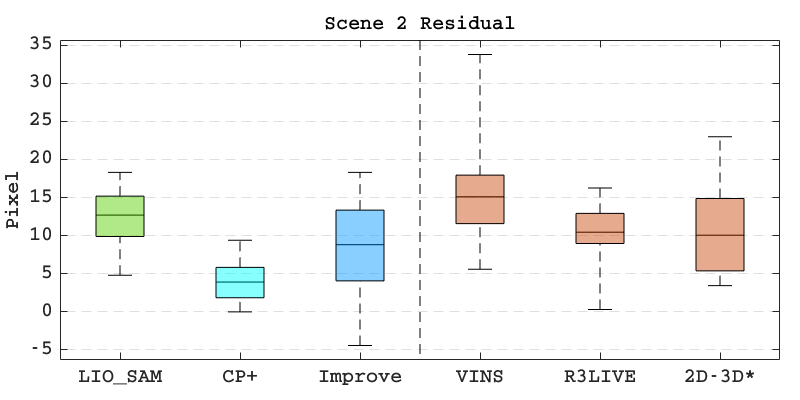}
\caption{The evaluation result with the device in Fig. \ref{fig:device:a}. 'LIO$\_$SAM' column is the performance of the LIO-SAM \cite{shan2020lio} algorithm, and 'Improve' denotes the improved performance of our LIO-SAM based CP$^+$ method. As comparison, we choose other three method, 'VINS' (VINS-MONO, \cite{qin2018vins}), 'R3LIVE' (R$^3$LIVE, \cite{lin2021r3live}) and '2D-3D' (\cite{feng20192d3d}).}
\label{fig:scene_12}
\end{figure}

Fig. \ref{fig:scene_12} shows the results of the comparison and that the performance of CP$^+$ is obviously better than the baseline method (LIO-SAM). Here the maximum number of iterations in our CP$^+$ method is set to three. The average computation time is $1.251s$.

\subsection{Augmentation for LIO\&VIO Algorithm}

\begin{figure}
\centering

\includegraphics[width=0.49\linewidth]{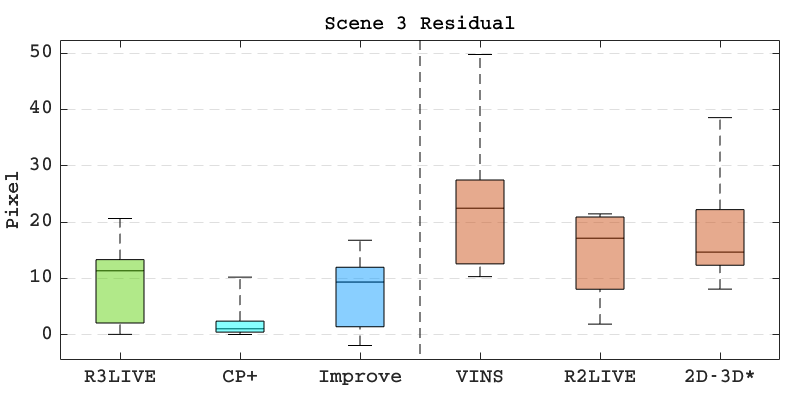}
\includegraphics[width=0.49\linewidth]{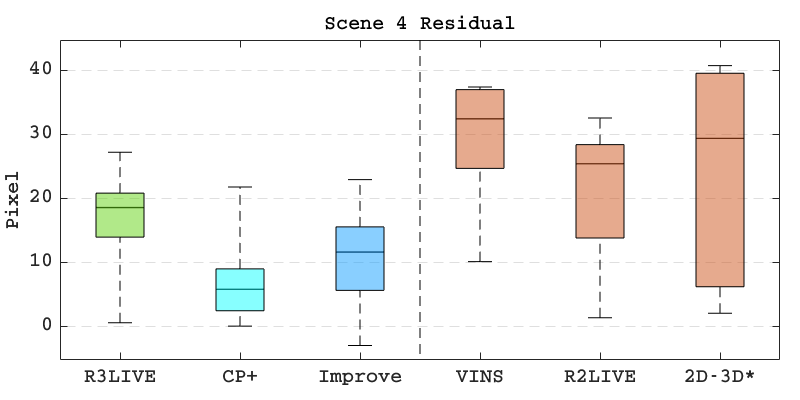}

\caption{The evaluation result with the device in Fig. \ref{fig:device:b}. Similar to Fig. \ref{fig:scene_12}, here 'Improve' denotes the improved performance of our R$^3$LIVE \cite{lin2021r3live} based CP$^+$ method, and 'R2LIVE' is the R$^2$LIVE \cite{lin2021r2live} method. }
\label{fig:scene_34}
\end{figure}

Similarly, here we run our method based on R$^3$LIVE \cite{lin2021r3live}. Fig. \ref{fig:scene_34} shows that our method can improve the pose estimates of LiDAR-Inertial-Visual systems. However, our method requires slightly higher precision for the initial pose estimates. Directly optimizing the results of VINS-mono \cite{qin2018vins}, the augmentation is often not obvious, not much different from 2D-3D matching methods introduced before.

\section{Conclusions}
In this paper, we presented a novel framework for camera localization augmentation with LiDAR maps, which integrates LiDAR and camera data through the 2D-3D correspondence detection and matching. Our novel framework employs dynamic object removal, ROI filtering and a multi-layer voxel map data structure to speed up the algorithm. The experimental evaluation of our approach showed its superior point cloud colorization performance when compared to state of the art approaches. Furthermore, the geometric registration of LiDAR and camera data in this paper can also provide new ideas for the joint bundle adjustment of these two kinds of data.

\bibliographystyle{plain}
\footnotesize 
\bibliography{main}

\begin{thebibliography}{10}

\bibitem{asvadi20163d}
Alireza Asvadi, Cristiano Premebida, Paulo Peixoto, and Urbano Nunes.
\newblock 3d lidar-based static and moving obstacle detection in driving
  environments: An approach based on voxels and multi-region ground planes.
\newblock {\em Robotics and Autonomous Systems}, 83:299--311, 2016.

\bibitem{campbell2018globally}
Dylan Campbell, Lars Petersson, Laurent Kneip, and Hongdong Li.
\newblock Globally-optimal inlier set maximisation for camera pose and
  correspondence estimation.
\newblock {\em IEEE transactions on pattern analysis and machine intelligence},
  42(2):328--342, 2018.

\bibitem{cattaneo2020cmrnet++}
Daniele Cattaneo, Domenico~Giorgio Sorrenti, and Abhinav Valada.
\newblock Cmrnet++: Map and camera agnostic monocular visual localization in
  lidar maps.
\newblock {\em arXiv preprint arXiv:2004.13795}, 2020.

\bibitem{chang2021hypermap}
Ming-Fang Chang, Joshua Mangelson, Michael Kaess, and Simon Lucey.
\newblock Hypermap: Compressed 3d map for monocular camera registration.
\newblock In {\em 2021 IEEE International Conference on Robotics and Automation
  (ICRA)}, pages 11739--11745. IEEE, 2021.

\bibitem{chen2019heterogeneous}
Hongyu Chen and S{\"o}ren Schwertfeger.
\newblock Heterogeneous multi-sensor calibration based on graph optimization.
\newblock In {\em 2019 IEEE International Conference on Real-time Computing and
  Robotics (RCAR)}, pages 158--163. IEEE, 2019.

\bibitem{chen2020advanced}
Hongyu Chen, Zhijie Yang, Xiting Zhao, Guangyuan Weng, Haochuan Wan, Jianwen
  Luo, Xiaoya Ye, Zehao Zhao, Zhenpeng He, Yongxia Shen, et~al.
\newblock Advanced mapping robot and high-resolution dataset.
\newblock {\em Robotics and Autonomous Systems}, 131:103559, 2020.

\bibitem{feng20192d3d}
Mengdan Feng, Sixing Hu, Marcelo~H Ang, and Gim~Hee Lee.
\newblock 2d3d-matchnet: Learning to match keypoints across 2d image and 3d
  point cloud.
\newblock In {\em 2019 International Conference on Robotics and Automation
  (ICRA)}, pages 4790--4796. IEEE, 2019.

\bibitem{fischler1981random}
Martin~A Fischler and Robert~C Bolles.
\newblock Random sample consensus: a paradigm for model fitting with
  applications to image analysis and automated cartography.
\newblock {\em Communications of the ACM}, 24(6):381--395, 1981.

\bibitem{haralick1994review}
Bert~M Haralick, Chung-Nan Lee, Karsten Ottenberg, and Michael N{\"o}lle.
\newblock Review and analysis of solutions of the three point perspective pose
  estimation problem.
\newblock {\em International journal of computer vision}, 13(3):331--356, 1994.

\bibitem{hornung2013octomap}
Armin Hornung, Kai~M Wurm, Maren Bennewitz, Cyrill Stachniss, and Wolfram
  Burgard.
\newblock Octomap: An efficient probabilistic 3d mapping framework based on
  octrees.
\newblock {\em Autonomous robots}, 34(3):189--206, 2013.

\bibitem{katz2015visibility}
Sagi Katz and Ayellet Tal.
\newblock On the visibility of point clouds.
\newblock In {\em Proceedings of the IEEE International Conference on Computer
  Vision}, pages 1350--1358, 2015.

\bibitem{katz2007direct}
Sagi Katz, Ayellet Tal, and Ronen Basri.
\newblock Direct visibility of point sets.
\newblock In {\em ACM SIGGRAPH 2007 papers}, pages 24--es. 2007.

\bibitem{kneip2014upnp}
Laurent Kneip, Hongdong Li, and Yongduek Seo.
\newblock Upnp: An optimal o (n) solution to the absolute pose problem with
  universal applicability.
\newblock In {\em European Conference on Computer Vision}, pages 127--142.
  Springer, 2014.

\bibitem{lepetit2009epnp}
Vincent Lepetit, Francesc Moreno-Noguer, and Pascal Fua.
\newblock Epnp: An accurate o (n) solution to the pnp problem.
\newblock {\em International journal of computer vision}, 81(2):155, 2009.

\bibitem{lin2021r3live}
Jiarong Lin and Fu~Zhang.
\newblock R3live: A robust, real-time, rgb-colored, lidar-inertial-visual
  tightly-coupled state estimation and mapping package.
\newblock {\em arXiv preprint arXiv:2109.07982}, 2021.

\bibitem{lin2021r2live}
Jiarong Lin, Chunran Zheng, Wei Xu, and Fu~Zhang.
\newblock R2live: A robust, real-time, lidar-inertial-visual tightly-coupled
  state estimator and mapping.
\newblock {\em IEEE Robotics and Automation Letters}, 6(4):7469--7476, 2021.

\bibitem{liu2021fast}
Xiyuan Liu, Chongjian Yuan, and Fu~Zhang.
\newblock Fast and accurate extrinsic calibration for multiple lidars and
  cameras.
\newblock {\em arXiv preprint arXiv:2109.06550}, 2021.

\bibitem{liu2021balm}
Zheng Liu and Fu~Zhang.
\newblock Balm: Bundle adjustment for lidar mapping.
\newblock {\em IEEE Robotics and Automation Letters}, 6(2):3184--3191, 2021.

\bibitem{qin2018vins}
Tong Qin, Peiliang Li, and Shaojie Shen.
\newblock Vins-mono: A robust and versatile monocular visual-inertial state
  estimator.
\newblock {\em IEEE Transactions on Robotics}, 34(4):1004--1020, 2018.

\bibitem{schauer2018peopleremover}
Johannes Schauer and Andreas N{\"u}chter.
\newblock The peopleremover—removing dynamic objects from 3-d point cloud
  data by traversing a voxel occupancy grid.
\newblock {\em IEEE robotics and automation letters}, 3(3):1679--1686, 2018.

\bibitem{schonberger2016structure}
Johannes~L Schonberger and Jan-Michael Frahm.
\newblock Structure-from-motion revisited.
\newblock In {\em Proceedings of the IEEE conference on computer vision and
  pattern recognition}, pages 4104--4113, 2016.

\bibitem{shan2020lio}
Tixiao Shan, Brendan Englot, Drew Meyers, Wei Wang, Carlo Ratti, and Daniela
  Rus.
\newblock Lio-sam: Tightly-coupled lidar inertial odometry via smoothing and
  mapping.
\newblock In {\em 2020 IEEE/RSJ International Conference on Intelligent Robots
  and Systems (IROS)}, pages 5135--5142. IEEE, 2020.

\bibitem{yu2020monocular}
Huai Yu, Weikun Zhen, Wen Yang, Ji~Zhang, and Sebastian Scherer.
\newblock Monocular camera localization in prior lidar maps with 2d-3d line
  correspondences.
\newblock In {\em 2020 IEEE/RSJ International Conference on Intelligent Robots
  and Systems (IROS)}, pages 4588--4594. IEEE, 2020.

\bibitem{yuan2021pixel}
Chongjian Yuan, Xiyuan Liu, Xiaoping Hong, and Fu~Zhang.
\newblock Pixel-level extrinsic self calibration of high resolution lidar and
  camera in targetless environments.
\newblock {\em IEEE Robotics and Automation Letters}, 6(4):7517--7524, 2021.

\bibitem{zhang2018tutorial}
Zichao Zhang and Davide Scaramuzza.
\newblock A tutorial on quantitative trajectory evaluation for visual
  (-inertial) odometry.
\newblock In {\em 2018 IEEE/RSJ International Conference on Intelligent Robots
  and Systems (IROS)}, pages 7244--7251. IEEE, 2018.

\end{thebibliography}

\end{document}